\documentclass[journal]{IEEEtran}
\IEEEoverridecommandlockouts
\usepackage{cite}
\usepackage{amsmath,amssymb,amsfonts}
\usepackage{algorithmic}
\usepackage{graphicx}
\usepackage{textcomp}
\usepackage{xcolor}
\usepackage{algorithm}
\usepackage{booktabs}
\usepackage{multirow}
\usepackage{makecell}
\usepackage{subfig}
\usepackage{url}
\def\BibTeX{{\rm B\kern-.05em{\sc i\kern-.025em b}\kern-.08em
    T\kern-.1667em\lower.7ex\hbox{E}\kern-.125emX}}

\begin{document}

\title{Cascading versus Joint Modeling for Hierarchical Offensive Language Detection}

\author{
	
Ruixing~Ren, Junhui~Zhao, Xiaoke~Sun

\thanks{ (Corresponding author: Junhui Zhao.)}%
\thanks{Ruixing Ren, Junhui Zhao are with the School of Electronic and Information Engineering, Beijing Jiaotong University, Beijing 100044, China.
	
Xiaoke Sun is with the National Computer Network Emergency Response Technical Team/Coordination Center of China (CNCERT/CC), Beijing 100029, China.
}
	
}

\maketitle

\begin{abstract}
Fine-grained offensive language detection organizes labels into a
hierarchical structure, for which two modeling paradigms exist:
cascaded decomposition and joint multi-task modeling. Prior work
rarely provides a direct, controlled comparison of the two paradigms
in terms of accuracy, parameter count, and inference latency, and
rarely verifies whether a chosen class-imbalance handling strategy is
actually optimal. This paper proposes a three-level cascaded
detection system whose training strategy is customized per subtask,
together with two verification mechanisms. First, a controlled
ablation study determines the best class-imbalance handling strategy
for each subtask. Second, a joint multi-task model with a shared
encoder is trained as an architectural control, yielding real
measurements along the dimensions of accuracy, parameter count, and
inference latency. Experiments show that the cascaded system attains
macro-F1 scores of 0.795, 0.716, and 0.557 on the three subtasks of
the official test set. The ablation study reveals that configuring
the loss function purely by imbalance-severity intuition is
suboptimal; reconfiguring based on the ablation results improves both
performance and stability. End-to-end cascade evaluation shows that
roughly one-fifth of the errors in the cascade pipeline originate
from the first-stage filter and cannot be corrected by subsequent
stages. Relative to the joint multi-task model, the cascaded
architecture achieves higher accuracy on all three subtasks, with a
7.1-point macro-F1 gain on the most severely imbalanced subtask, at
the cost of three times the parameters and 1.67 times the inference
latency. Together, these results establish an explicit, quantifiable
trade-off between the accuracy advantage of cascaded architectures
and their deployment cost.
\end{abstract}


\section{Introduction}\label{Sec1}

Automatic detection of offensive content on social media platforms
remains a core technical challenge for content safety
governance~\cite{RenDCAN}. Detection methods in this area have
progressively moved from traditional machine learning to fine-tuning
pretrained language models such as BERT, yet research remains
heavily concentrated on English, and fine-grained, multi-level
offensive-language judgments receive markedly less systematic
validation than coarse-grained binary
classification~\cite{ramos2024review}. Compared with a simple binary
judgment, fine-grained detection, that is, jointly determining
whether content is offensive, whether the offense targets an
identifiable entity, and whether the target is an individual or a
group, can provide moderators with more actionable
guidance~\cite{TOXIN,COLD}. Such fine-grained label schemes naturally
form a hierarchical structure, and how to choose an appropriate
modeling paradigm for this structure, as well as how to handle the
increasingly severe class imbalance found deeper in the hierarchy,
remains an underexplored engineering problem.

On the task side, the OLID dataset and its three-level annotation
scheme, introduced by Zampieri et
al.~\cite{zampieri2019olid,zampieri2019semeval}, are among the most
widely used benchmarks in this direction, and a later multilingual
extension~\cite{zampieri2020semeval} further confirmed the generality
of this annotation framework. Around similar fine-grained offensive
content judgment tasks, recent systems have continued to combine
ensembling with fine-tuning. Ganguly et
al.~\cite{ganguly2024masonperplexity} used XLM-RoBERTa together with
a multi-model ensemble in the 2024 multimodal hate speech detection
task, ranking third on both the identification and the target
localization subtasks; Guragain et
al.~\cite{guragain2025nlpineers} ensembled several multilingual BERT
variants for Devanagari hate speech detection in 2025 and applied
back-translation augmentation specifically to address class
imbalance. All of these systems are designed for a single hierarchy
level and do not address the choice of modeling paradigm for the
hierarchy itself.

Two modeling paradigms exist for hierarchical label structures, a
distinction formalized in the survey of hierarchical classification by
Silla and Freitas~\cite{silla2011hierarchical}: local classifier
approaches, corresponding to cascaded decomposition, train a separate
classifier for each level, where the prediction at one level
determines whether a sample proceeds to the next, while global
classifier approaches, corresponding to joint multi-task modeling,
instead use a single model to predict all levels simultaneously. Dai et
al.~\cite{dai2020kungfupanda} proposed a shared-encoder joint
multi-task method for the same three-level OLID task addressed in
this paper, exploiting mutually reinforcing supervision across
subtasks to achieve results close to the best system of that year
using only OLID~\cite{zampieri2019olid} data. Mnassri et
al.~\cite{mnassri2023emotion} combined joint multi-task learning with
external emotion features, using BERT and multilingual BERT to
address offensive content detection in monolingual and cross-lingual
settings respectively, alleviating the imbalance and scarcity of
labeled data. Saha et al.~\cite{saha2025retriv} further extended
joint modeling to the simultaneous prediction of type, target, and
severity in the 2025 Bangla hate speech identification task, a
structure closely resembling the three-level OLID scheme. These works
all validate the effectiveness of joint modeling, but the real
trade-off between joint modeling and cascaded decomposition in terms
of accuracy, parameter count, and inference latency has not
previously been directly examined.

At the same time, deeper levels of the hierarchy tend to have fewer
samples and more severe class imbalance; the second level of the
OLID three-level scheme already reaches an imbalance ratio of 7.5 to
1. Class imbalance is a long-studied problem in machine learning, and
Johnson and Khoshgoftaar~\cite{johnson2019survey} survey the range of
deep-learning remedies, which broadly fall into resampling the
training distribution, exemplified by the SMOTE oversampling method
of Chawla et al.~\cite{chawla2002smote}, and reweighting the loss
contribution of each sample, as in the Focal Loss proposed by Lin et
al.~\cite{lin2017focal}. The system of Guragain et
al.~\cite{guragain2025nlpineers} mentioned above likewise designs a
dedicated augmentation scheme for this problem. However, how to
choose an appropriate strategy for different levels and different
degrees of imbalance is, in most prior work, left to intuition about
the imbalance ratio rather than being verified through systematic
ablation.

These three gaps, namely the absence of a genuine, multi-dimensional
comparison between hierarchical modeling paradigms, the absence of
systematic verification of class-imbalance handling strategies, and
the absence of a quantitative analysis of error propagation through
the hierarchy, together motivate this work. Table~\ref{tab:comparison}
compares representative prior work with this paper along five
dimensions.

\begin{table*}[t]
\centering
\caption{Comparison with Representative Prior Work}
\label{tab:comparison}
\renewcommand{\arraystretch}{1.2}
\begin{tabular}{lccccc}
\toprule
Work & Cascade & Joint Comparison & Loss Ablation & E2E Error & Params/Latency \\
\midrule
Zampieri et al.~\cite{zampieri2019olid,zampieri2019semeval} & Hierarchical protocol & $\times$ & $\times$ & $\times$ & $\times$ \\
Dai et al.~\cite{dai2020kungfupanda} & $\times$ & Joint model only & $\times$ & $\times$ & $\times$ \\
Mnassri et al.~\cite{mnassri2023emotion} & $\times$ & Joint model only & $\times$ & $\times$ & $\times$ \\
Saha et al.~\cite{saha2025retriv} & $\times$ & Joint model only & $\times$ & $\times$ & $\times$ \\
This work & \checkmark & \checkmark{} & \checkmark & \checkmark & \checkmark \\
\bottomrule
\end{tabular}
\end{table*}

To address this gap, this paper proposes a three-level cascaded
offensive language detection system whose training strategy is
customized per subtask, and systematically validates the
effectiveness of its design choices. The contributions of this paper
are summarized as follows.

\begin{itemize}
  \item A task-adaptive class-imbalance handling scheme. The loss
    function is selected independently for each subtask, and the
    class weighting of Focal Loss is vectorized rather than left as a
    scalar; combined with a controlled-variable ablation protocol,
    this corrects the practice of configuring the strategy purely by
    imbalance-severity intuition.
  \item A joint multi-task control model with task gating. A gating
    term that depends on the ground-truth label of the upper level
    masks out invalid loss contributions, making the cascaded
    architecture and the joint model comparable under identical
    experimental conditions and yielding real measurements along the
    dimensions of accuracy, parameter count, and inference latency.
  \item A two-tier evaluation protocol. Beyond the existing per-task
    standard evaluation, an end-to-end cascade evaluation is
    introduced, in which the model's own predictions are propagated
    level by level, attributing the system's final errors to a
    specific subtask stage.
\end{itemize}

The remainder of this paper is organized as follows. Section~\ref{Sec2}
describes the data, model architecture, and experimental design.
Section~\ref{Sec3} presents the experimental results and analysis.
Section~\ref{Sec4} concludes the paper and discusses its limitations.

\section{Method}\label{Sec2}

\subsection{Method Overview}\label{Sec2.0}

The proposed offensive language detection system consists of three
components that work together, with the data flow and training
pipeline shown in Fig.~\ref{fig:framework}. At the data level, input
text is organized under the OLID three-level annotation scheme
(Section~\ref{Sec2.1}) into three hierarchically dependent subtasks:
determining offensiveness, determining whether the offense targets an
identifiable entity, and determining the type of target. At the model
level, each subtask has an independent classification head that
takes the semantic representation produced by the shared backbone
(Section~\ref{Sec2.2}) and, as described in Section~\ref{Sec2.3},
maps it to a class probability distribution; this distribution then
serves as the input to the task-adaptive loss function of
Section~\ref{Sec2.4}, which drives the gradient updates of the
classification head, with the specific form of the loss function for
each subtask determined by the ablation protocol of
Section~\ref{Sec2.7}. As a quantitative control, Section~\ref{Sec2.5}
introduces a joint multi-task model with task gating that keeps every
other setting fixed and varies only the architecture, so that any
difference in accuracy can be explicitly attributed to the modeling
paradigm. Section~\ref{Sec2.6} gives the training details shared by
all models. Finally, the two-tier evaluation protocol of
Section~\ref{Sec2.8} quantifies system performance from both the
standard benchmark perspective and the end-to-end cascade
perspective.

\subsection{Data and the Three-Level Label Scheme}\label{Sec2.1}

This paper uses the OLID~\cite{zampieri2019olid} dataset, which after
cleaning contains 13,203 English tweets, split 9:1 into training and
validation sets of 11,882 and 1,321 examples respectively, with a
separate official test set used for final reporting containing 860,
240, and 213 examples for subtasks A, B, and C. The dataset follows a
three-level annotation scheme: subtask A determines offensiveness,
with labels OFF or NOT, at 33.3\% and 66.7\% of the training set;
subtask B is annotated only for OFF examples and determines whether
the offense targets an identifiable entity, with labels TIN or UNT at
88.2\% and 11.8\%; subtask C is annotated only for TIN examples and
determines the type of target, with labels IND, GRP, or OTH,
representing an individual, a group, or other, at 62.3\%, 27.3\%, and
10.3\% respectively. The imbalance ratios of the three subtasks are
approximately 2 to 1, 7.5 to 1, and 6 to 1, all different from one
another, and this difference is the direct motivation for customizing
the training strategy per task in this paper.

\subsection{Backbone Model}\label{Sec2.2}

This paper uses \texttt{bert-base-uncased}~\cite{devlin2019bert} as
the shared semantic backbone, encoding the input text into a
fixed-dimensional contextual representation; the classification heads
of the three subtasks are trained independently. The backbone does
not participate in the hierarchical decision logic described in
Section~\ref{Sec2.1} and only supplies a unified semantic feature for
each classification head to use; the classification head structure
and probability mapping are given in Section~\ref{Sec2.3}.

\subsection{Cascaded Classification Heads and Probability Computation}\label{Sec2.3}

The three subtasks share the same classification head structure,
whose input is the \texttt{[CLS]} representation $h \in
\mathbb{R}^{768}$ produced by the backbone described in
Section~\ref{Sec2.2}, and whose output is the class probability
distribution for that subtask. The classification head first applies
Dropout~\cite{srivastava2014dropout} and then a fully connected layer, mapping $h$ to a raw score
vector $z$ of length $K$, which is then converted by the Softmax
function into a class probability distribution
$\mathbf{p}=[p_1,\dots,p_K]$:
\begin{equation}
z = W h + b, \qquad p_i = \frac{e^{z_i}}{\sum_{j=1}^{K} e^{z_j}}
\label{eq:softmax}
\end{equation}
where $K$ is the number of classes for that subtask, equal to 2 for
subtasks A and B and 3 for subtask C. This probability distribution
$\mathbf{p}$ serves two purposes. During training, it is the input to
the loss function described in Section~\ref{Sec2.4}, which measures
the deviation between the prediction and the ground truth and drives
the gradient update of the model parameters; during inference, the
class index attaining the maximum value is taken as the prediction
for that subtask and determines, under the cascaded structure,
whether the next subtask is invoked. The full cascade inference
procedure is given in Algorithm~\ref{alg:cascade}, where
$\text{argmax}(\cdot)$ returns the class index with the highest
probability and $M_A$, $M_B$, $M_C$ denote the trained models for the
three subtasks.

\begin{algorithm}[t]
\caption{Three-Level Cascade Inference}
\label{alg:cascade}
\begin{algorithmic}[1]
\REQUIRE Input text $x$; subtask models $M_A, M_B, M_C$
\ENSURE Three-level prediction $(\hat{y}_A, \hat{y}_B, \hat{y}_C)$
\STATE $\mathbf{p}_A \leftarrow M_A(x)$
\STATE $\hat{y}_A \leftarrow \text{argmax}(\mathbf{p}_A)$
\STATE $\hat{y}_B \leftarrow \varnothing;\ \hat{y}_C \leftarrow \varnothing$
\IF{$\hat{y}_A = \text{OFF}$}
    \STATE $\mathbf{p}_B \leftarrow M_B(x)$
    \STATE $\hat{y}_B \leftarrow \text{argmax}(\mathbf{p}_B)$
    \IF{$\hat{y}_B = \text{TIN}$}
        \STATE $\mathbf{p}_C \leftarrow M_C(x)$
        \STATE $\hat{y}_C \leftarrow \text{argmax}(\mathbf{p}_C)$
    \ENDIF
\ENDIF
\RETURN $(\hat{y}_A, \hat{y}_B, \hat{y}_C)$
\end{algorithmic}
\end{algorithm}

Algorithm~\ref{alg:cascade} explicitly reflects two properties of the
cascaded architecture proposed in this paper. First, the number of
models a sample passes through adapts to its content complexity; a
NOT example needs only a single forward pass to return a result,
which lowers average inference latency. Second, a prediction error at
an earlier stage cannot be corrected by a later stage: if
$\hat{y}_A$ misclassifies a truly OFF example as NOT, $M_B$ and $M_C$
are never invoked, and that example permanently loses any chance of
correction on subtasks B and C. This property directly motivates the
end-to-end cascade evaluation protocol of Section~\ref{Sec2.8}.

\subsection{Task-Adaptive Class-Imbalance Optimization}\label{Sec2.4}

The imbalance ratios of the three subtasks shown in
Section~\ref{Sec2.1} are approximately 2 to 1, 7.5 to 1, and 6 to 1,
all different from one another; this difference means that a single
loss function is unlikely to be simultaneously optimal for all three
subtasks, and an appropriate class-imbalance handling strategy must
be chosen for each. This paper considers two representative loss
functions, which share the same inverse-frequency class weighting and
differ only in whether they include a hard-example focusing
mechanism.

Given the class probability $\mathbf{p}$ output by the classification
head via Eq.~\eqref{eq:softmax}, class-weighted cross-entropy
reweights the contribution of each class by the inverse of its
frequency in the training set, mitigating the extent to which the
majority class dominates the direction of the gradient update:
\begin{equation}
\mathcal{L}_{\text{CE}} = -\sum_{c=1}^{K} w_c\, y_c \log p_c, \qquad
w_c = \frac{N}{K \times n_c}
\label{eq:weighted_ce}
\end{equation}
where $N$ is the total number of training samples, $y_c$ is the
one-hot encoding of the ground-truth label, $n_c$ is the number of
samples in class $c$, and $w_c$ is the weight assigned to class $c$.
Equation~\eqref{eq:weighted_ce} adjusts only the overall contribution
of each class and treats easy and hard examples within the same
class identically. For more severely imbalanced settings, Focal
Loss~\cite{lin2017focal} builds on this by introducing a hard-example
focusing mechanism that further downweights easy examples:
\begin{equation}
\mathcal{L}_{\text{focal}} = -\alpha_y (1 - p_t)^{\gamma} \log p_t
\label{eq:focal}
\end{equation}
where $p_t$ is the model's predicted probability for the true class
and $\gamma$ is the focusing parameter, set to 2.0 in this paper. To
let Eq.~\eqref{eq:focal} share the same class-weighting abstraction
as Eq.~\eqref{eq:weighted_ce} and thereby support a rigorous
comparison, this paper sets $\alpha_y$ to the same per-class weight
vector as $w_c$ in Eq.~\eqref{eq:weighted_ce}, rather than letting it
degenerate into a global scalar; only in this form does the loss
function simultaneously provide class reweighting and hard-example
focusing. Which loss function each subtask should use is not
specified directly by its imbalance ratio, but is instead determined
independently for each subtask through the ablation protocol of
Section~\ref{Sec2.7}, with results reported in Section~\ref{Sec3.3}.

\begin{figure*}[t]
	\centering
	\includegraphics[width=0.9\linewidth]{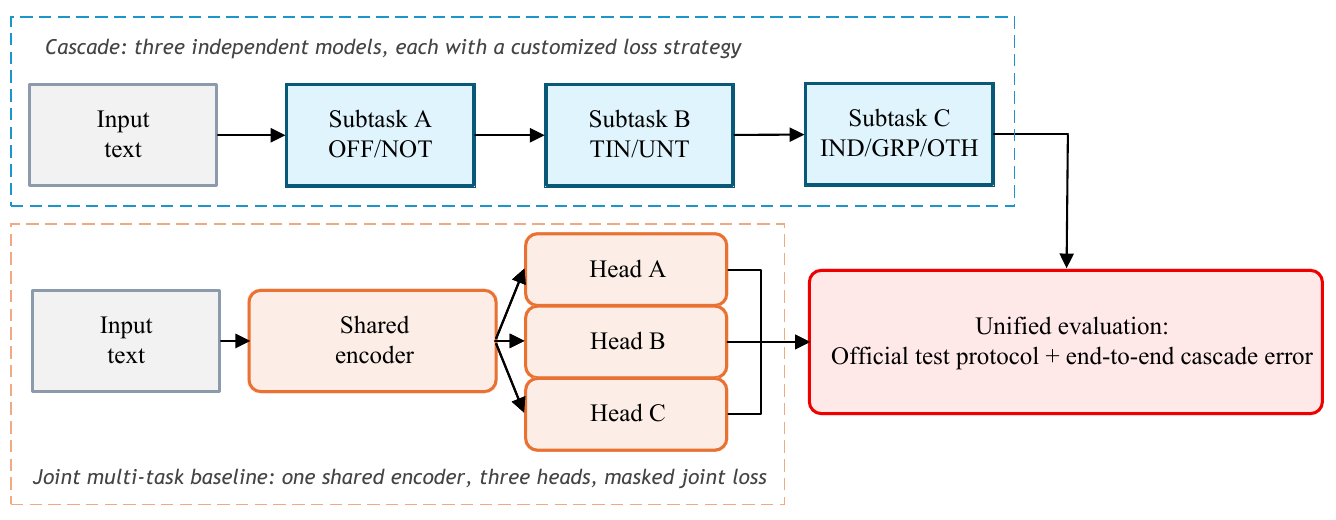}
	\caption{The cascaded architecture (top) and the joint multi-task
		control model (bottom), sharing a unified evaluation protocol.}
	\label{fig:framework}
\end{figure*}
\subsection{Joint Multi-Task Control Model}\label{Sec2.5}

To directly compare the cascaded architecture of
Section~\ref{Sec2.3} with the joint modeling paradigm under identical
experimental conditions, this paper constructs a joint multi-task
model that differs from the cascaded architecture only at the
architectural level. Its encoder is shared across the three subtasks
while the classification heads remain independent; the input text is
encoded once by the shared encoder into a single \texttt{[CLS]}
representation, which is then fed to all three classification heads
to obtain the class probabilities for all three subtasks. Because a
lower-level label is available for a given training example only when
the corresponding hierarchical dependency is satisfied, subtask B
only when the subtask A label is OFF, and subtask C only when the
subtask B label is TIN, the three loss terms cannot be computed
uniformly for every example. To address this, a task-gating term
that depends on the ground-truth label of the upper level is
introduced into the joint loss to mask out inapplicable loss
contributions:
\begin{equation}
\mathcal{L}_{\text{joint}} = \mathcal{L}_A + \mathcal{L}_B \cdot m_B + \mathcal{L}_C \cdot m_C
\label{eq:joint_loss}
\end{equation}
where $\mathcal{L}_A$, $\mathcal{L}_B$, $\mathcal{L}_C$ are each
computed using the same loss function described in
Section~\ref{Sec2.4}, $m_B$ equals 1 when the subtask A label of the
example is OFF and 0 otherwise, and $m_C$ equals 1 when the subtask B
label is TIN and 0 otherwise. This gating design preserves the
representation-sharing effect of the shared encoder while avoiding
the noise that would be introduced by driving gradient updates with
invalid labels. This joint model uses the same backbone, the same
training data, and the same evaluation protocol as the cascaded
architecture of Section~\ref{Sec2.3}; the only difference is that the
three tasks share a single encoder rather than being trained
independently, so that the differences in accuracy, parameter count,
and inference latency observed in Section~\ref{Sec3.5} can be
attributed directly to the modeling paradigm itself.

\subsection{Optimizer and Training Setup}\label{Sec2.6}

This paper follows the general fine-tuning recipe for BERT-based text
classification established by Sun et al.~\cite{sun2019finetune}.
Training uses the AdamW optimizer~\cite{loshchilov2019adamw} with a
discriminative learning rate~\cite{howard2018ulmfit}, $2\times10^{-5}$
for
the encoder and $1\times10^{-4}$ for the classification heads,
together with a schedule that linearly warms up over the first 10\%
of training steps and then decays linearly. Training length is
controlled by early stopping with a patience of 3, monitored on
validation macro-F1, with a batch size of 16, a maximum sequence
length of 128, and FP16 mixed precision enabled throughout training.

\subsection{Ablation Design}\label{Sec2.7}

To verify the choice of class-imbalance handling strategy, this paper
tests three loss functions on each of the three subtasks: plain
cross-entropy, the class-weighted cross-entropy of
Eq.~\eqref{eq:weighted_ce}, and the Focal Loss of
Eq.~\eqref{eq:focal}, giving $3 \times 3 = 9$ configurations. Each
configuration is independently repeated 3 times, with the random seed
reset before every repetition, and the mean and standard deviation
are reported to avoid a single run's outcome misleadingly driving the
conclusion.

\subsection{Evaluation Protocol}\label{Sec2.8}

This paper adopts two complementary evaluation protocols. The
per-task standard evaluation is carried out on each subtask's own
official test set and is directly comparable with results reported in
prior work. The end-to-end cascade evaluation instead follows the
inference procedure of Algorithm~\ref{alg:cascade} on the validation
set, propagating the model's own predictions, rather than the ground
truth, through subtasks A, B, and C in sequence, and counts an
example as end-to-end correct only when all three predicted levels
match the ground truth; this quantifies the real impact of error
propagation through the cascaded structure described in
Section~\ref{Sec2.3}. This protocol uses the validation set rather
than the official test set because the official test set is released
separately per subtask, so no public subset of examples carries
ground-truth labels for all three levels simultaneously.

Because all three subtasks exhibit some degree of class imbalance,
plain accuracy would be dominated by the majority class and fail to
reflect the model's ability to discriminate minority classes; both
protocols therefore use macro-F1, which weighs the performance on
each class equally, as the core metric:
\begin{equation}
\text{F1}_{\text{macro}} = \frac{1}{K}\sum_{i=1}^{K}
\frac{2\,\text{Precision}_i \cdot \text{Recall}_i}{\text{Precision}_i + \text{Recall}_i}
\label{eq:macrof1}
\end{equation}
where $K$ is the number of classes in the task being evaluated, and
$\text{Precision}_i$ and $\text{Recall}_i$ are the precision and
recall for class $i$.

\section{Experiments and Results}\label{Sec3}

\subsection{Experimental Setup}\label{Sec3.1}

Experiments were carried out on a single NVIDIA RTX 4060 Ti with
16GB of memory, running PyTorch 2.3.0+cu121. Fig.~\ref{fig:framework}
shows the structure of the cascaded architecture and the joint
multi-task control model, which share the same evaluation protocol to
ensure strictly comparable results.

\begin{figure}[t]
\centering
\includegraphics[width=\linewidth]{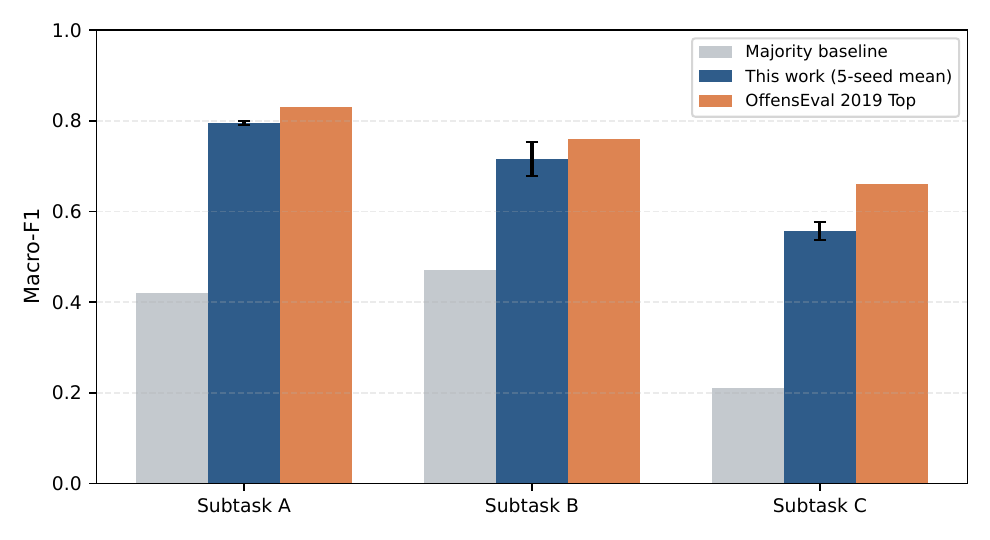}
\caption{Macro-F1 of the three subtasks on the official test set,
compared with the majority-class baseline and the best official
OffensEval 2019 system; error bars show the standard deviation over 5
independent repetitions.}
\label{fig:main_results}
\end{figure}

Table~\ref{tab:main_results} and Fig.~\ref{fig:main_results} report
the final results of the cascaded system on the official test set,
given as the mean and standard deviation over 5 independent
repetitions, each using a different random seed reset before
training so that the results are mutually independent. The macro-F1
scores of subtasks A, B, and C are 0.795, 0.716, and 0.557
respectively, all significantly above the majority-class baselines of
0.42, 0.47, and 0.21; subtasks A and B are on the same order as the
0.83 and 0.76 achieved by the best official systems of OffensEval
2019~\cite{zampieri2019semeval}, while subtask C still trails the
0.66 achieved by the best official system, which is directly related
to this subtask having the fewest training examples and the hardest
three-way decision boundary. The standard deviation for subtask B
reaches 0.037, markedly larger than the 0.004 for subtask A, which is
directly related to subtask B having both the smallest official test
set, only 240 examples, and the most severe class imbalance; small
test sets are more sensitive to individual prediction errors.

\begin{table}[t]
\centering
\caption{Main Results: Macro-F1 on the Official Test Set, Mean $\pm$ Standard Deviation over 5 Independent Repetitions}
\label{tab:main_results}
\renewcommand{\arraystretch}{1.2}
\begin{tabular}{lccc}
\toprule
 & Subtask A & Subtask B & Subtask C \\
\midrule
Majority baseline & 0.42 & 0.47 & 0.21 \\
This work & 0.795$\pm$0.004 & 0.716$\pm$0.037 & 0.557$\pm$0.020 \\
OffensEval Top & 0.83 & 0.76 & 0.66 \\
\bottomrule
\end{tabular}
\end{table}

\begin{figure*}[t]
\centering
\includegraphics[width=\linewidth]{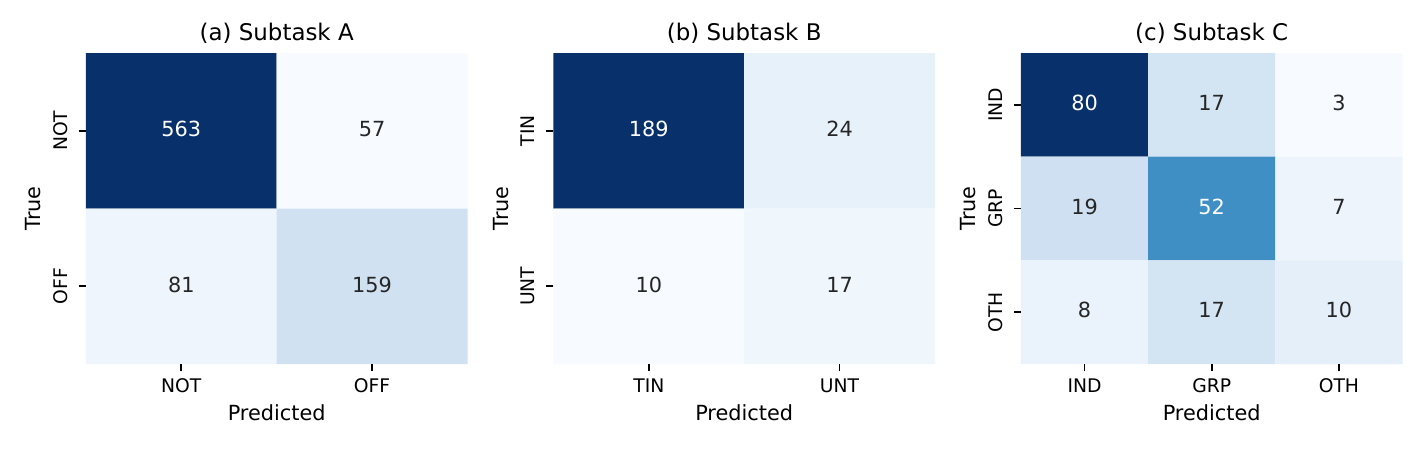}
\caption{Confusion matrices of the three subtasks on the official test set.}
\label{fig:confusion}
\end{figure*}

Fig.~\ref{fig:confusion} shows the confusion matrices of the three
subtasks on the official test set, taken from a representative
repetition that attains the mean level reported in
Table~\ref{tab:main_results}. Errors on subtask A are concentrated in
the direction of offensive examples being misclassified as
non-offensive; errors on subtask C mainly occur between GRP and the
other two classes, indicating that the boundary between group targets
and individual or other targets is more easily confused at the level
of surface text features.

\subsection{Loss-Strategy Ablation}\label{Sec3.3}

\begin{figure}[t]
\centering
\includegraphics[width=\linewidth]{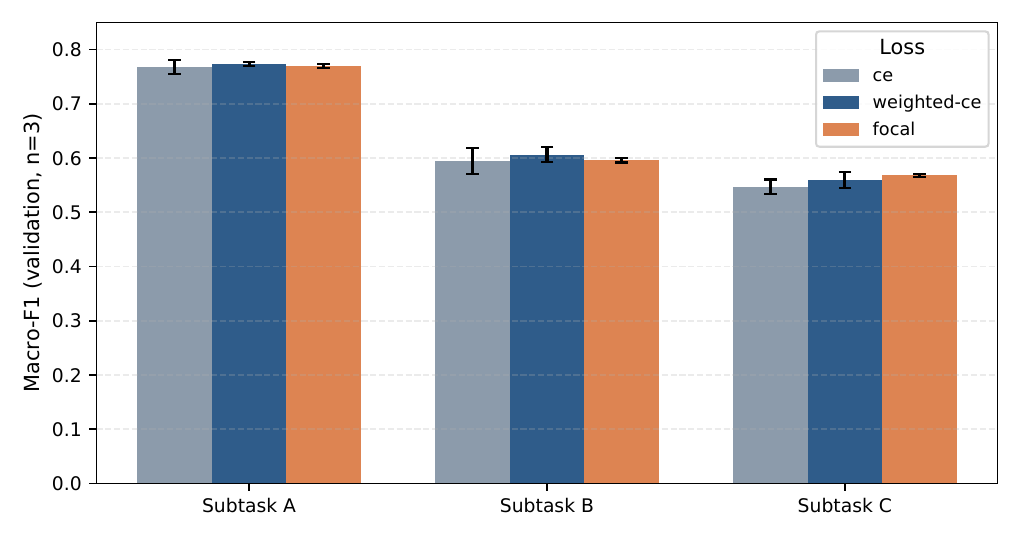}
\caption{Validation macro-F1 of the three loss-function strategies on
the three subtasks; error bars show the standard deviation over 3
independent repetitions.}
\label{fig:ablation}
\end{figure}

Table~\ref{tab:ablation} and Fig.~\ref{fig:ablation} report the
ablation results. The results show that the intuition-based strategy
of assigning a simpler loss to a less-imbalanced subtask and a more
complex loss to a more-imbalanced one does not hold: the optimal
strategy for each of the three subtasks disagrees with this
intuition-based configuration. The best strategy for both subtasks A
and B is class-weighted cross-entropy, improving over the
intuition-based configuration by 0.6 and 1.2 points respectively while
also reducing the standard deviation; the best strategy for subtask C
is Focal Loss, improving by 0.9 points and reducing the standard
deviation to one-fifth of its previous value. These results indicate
that the choice of class-imbalance handling strategy must be verified
experimentally for each specific task rather than inferred from the
imbalance ratio alone.

\begin{table}[t]
\centering
\caption{Loss-Strategy Ablation Results: Validation Macro-F1, Mean $\pm$ Standard Deviation, $n=3$}
\label{tab:ablation}
\renewcommand{\arraystretch}{1.2}
\begin{tabular}{lccc}
\toprule
 & Cross-Entropy & Class-Weighted & Focal \\
\midrule
Subtask A & 0.768$\pm$0.013 & \textbf{0.774$\pm$0.004} & 0.770$\pm$0.004 \\
Subtask B & 0.594$\pm$0.024 & \textbf{0.606$\pm$0.014} & 0.596$\pm$0.004 \\
Subtask C & 0.547$\pm$0.013 & 0.559$\pm$0.015 & \textbf{0.568$\pm$0.003} \\
\bottomrule
\end{tabular}
\end{table}

To ensure the statistical comparability and reliability of the
ablation study above, every independent run used an independently
reset random seed within an isolated run context, preventing the
order in which different configurations consume randomness from
interfering across configurations\footnote{In an early pilot version,
a random seed fixed only at the outer loop caused the random state to
drift cumulatively with the order of configurations, producing a
cross-configuration bias that made the baselines incomparable; this
issue has been fixed through the isolation strategy described above.}.
The final results in Table~\ref{tab:ablation} and
Table~\ref{tab:main_results} were both measured under this strict
isolation condition. Based on the ablation results in
Table~\ref{tab:ablation}, this paper adopts the configuration
identified as optimal, namely class-weighted cross-entropy for
subtasks A and B and Focal Loss for subtask C, as the training
configuration used to report official test set performance in
Table~\ref{tab:main_results}.

\subsection{End-to-End Cascade Error Propagation}\label{Sec3.4}

An end-to-end cascade evaluation was carried out on the validation
set of 1,321 examples, in which the model's own predictions, rather
than the ground truth, are propagated through subtasks A, B, and C in
sequence. The end-to-end strict accuracy, requiring all three
predicted levels to match the ground truth, is 0.717, with errors
originating from subtask A accounting for 19.6\% of all errors. This
proportion indicates that roughly one-fifth of the final errors in
the cascaded structure already occur at the first-stage filter; no
matter how much the discriminative accuracy of subtasks B and C is
improved, these errors cannot be corrected, and they constitute an
upper bound on the error propagation inherent to the cascaded
architecture itself.

\subsection{Cascaded Architecture versus Joint Multi-Task Model}\label{Sec3.5}

\begin{figure*}[t]
\centering
\includegraphics[width=0.9\linewidth]{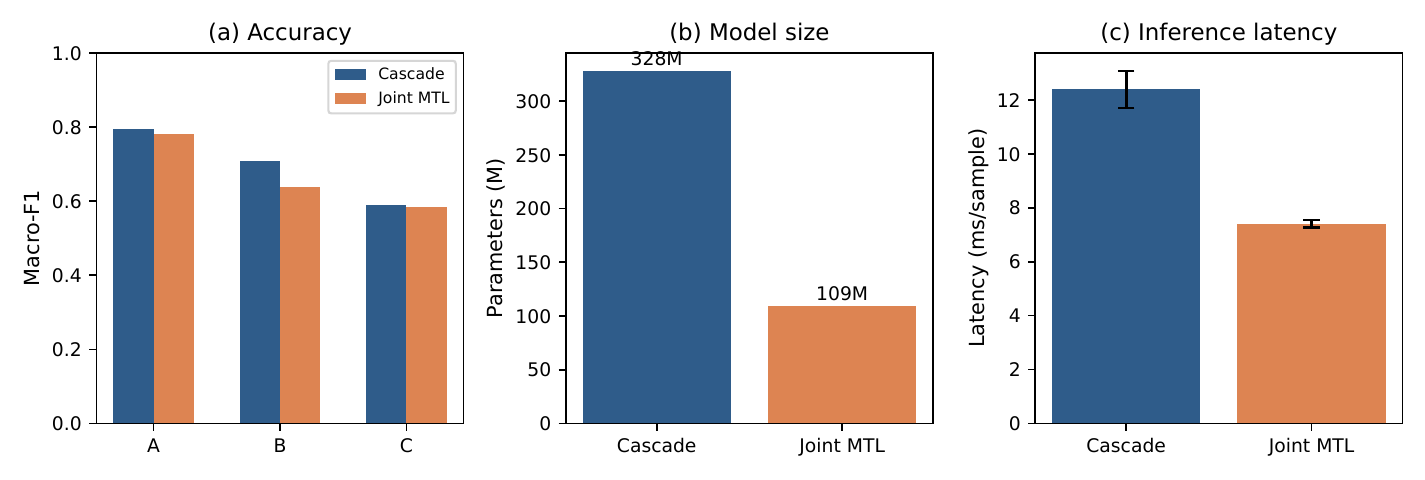}
\caption{Comparison between the cascaded architecture and the joint
multi-task model along accuracy, parameter count, and inference
latency.}
\label{fig:cascade_vs_joint}
\end{figure*}

Table~\ref{tab:cascade_vs_joint} and Fig.~\ref{fig:cascade_vs_joint}
give a direct comparison, on the official test set, between the
cascaded architecture and the joint multi-task model of
Section~\ref{Sec2.5}. The cascaded architecture achieves higher
accuracy on all three subtasks; the gains on subtasks A and C are
smaller, 1.4 and 0.7 points respectively, while the gain on subtask B
reaches 7.1 points, the largest of the three. This difference is
directly related to subtask B having the most severe class imbalance:
the cascaded architecture assigns this subtask an independent model
with a dedicated class-weighted loss that can be optimized entirely
for this one task, whereas the shared encoder of the joint model must
balance the gradient signals of all three tasks at once, so the
negative-transfer effect of representation sharing is most pronounced
on the task with the most severe imbalance and the greatest learning
difficulty.

\begin{table}[t]
\centering
\caption{Cascaded Architecture versus Joint Multi-Task Model on the Official Test Set}
\label{tab:cascade_vs_joint}
\renewcommand{\arraystretch}{1.2}
\begin{tabular}{lccc}
\toprule
 & Cascade & Joint MTL & Difference \\
\midrule
Subtask A Macro-F1 & 0.794 & 0.780 & +1.4pp \\
Subtask B Macro-F1 & 0.709 & 0.638 & +7.1pp \\
Subtask C Macro-F1 & 0.590 & 0.583 & +0.7pp \\
Parameters & 328.5M & 109.5M & 3.00$\times$ \\
Inference latency & 12.40$\pm$0.69ms & 7.41$\pm$0.14ms & 1.67$\times$ \\
\bottomrule
\end{tabular}
\end{table}

\begin{figure*}[t]
\centering
\includegraphics[width=0.8\linewidth]{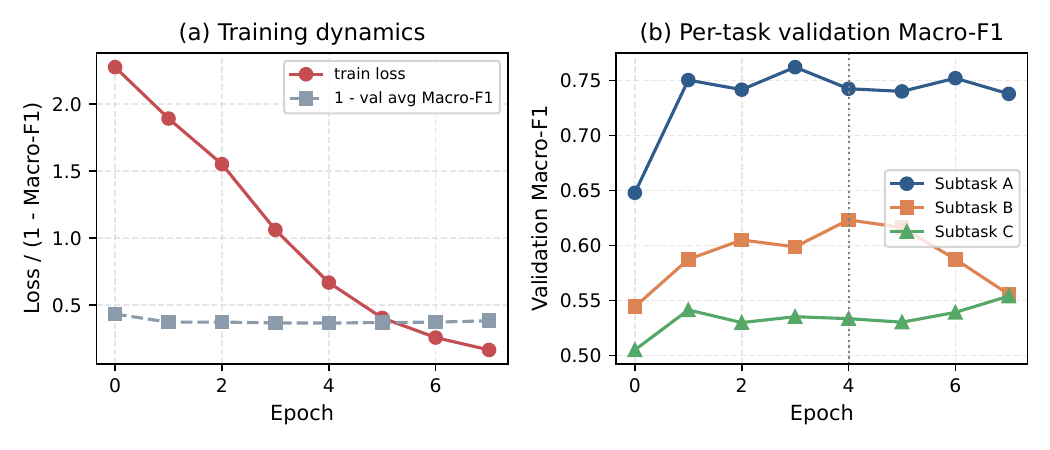}
\caption{Training dynamics of the joint multi-task model. Subplot (a)
shows overall training dynamics; subplot (b) shows the validation
macro-F1 of each of the three subtasks.}
\label{fig:joint_training}
\end{figure*}

Fig.~\ref{fig:joint_training} shows the training dynamics of the
joint multi-task model. Subplot (a) shows that while the training loss
continues to decrease, the combined validation metric peaks around
epoch 4 and then declines, a typical overfitting signature; subplot
(b) breaks this down by subtask and shows that the validation curve
for subtask B fluctuates the most, corroborating the result in
Table~\ref{tab:cascade_vs_joint} that this subtask suffers the
largest accuracy loss under the joint model: on the subtask with the
most severe class imbalance, the shared encoder not only attains
lower final accuracy but is also less stable during training.

Taking the three dimensions of Table~\ref{tab:cascade_vs_joint}
together, there is no absolute winner between the cascaded
architecture and the joint multi-task model; rather, they present a
quantifiable engineering trade-off. If a deployment scenario is
highly sensitive to parameter count and inference latency, the joint
multi-task model trades a modest accuracy cost for a one-third
reduction in parameters and nearly a 40\% reduction in latency. If a
deployment scenario instead prioritizes detection accuracy on
class-imbalanced subtasks, the additional deployment cost of the
cascaded architecture buys a meaningful accuracy gain, with the
largest benefit on severely imbalanced subtasks such as subtask B.

\subsection{Limitations}\label{Sec3.6}

The loss-strategy ablation in this paper uses 3 independent
repetitions, a relatively limited sample; the optimal strategy for
subtask C is determined mainly from validation set results, and the
official test set, which is small, only 213 examples with the
smallest class OTH containing only 35 examples, showed observations
under some configurations that were not fully consistent with the
validation set, which is reported honestly in the text. The
comparison between the cascaded architecture and the joint
multi-task model is currently based on a single training run for
each, without multiple repetitions to obtain confidence intervals,
which is a statistical limitation of this paper left for future work.

\section{Conclusion}\label{Sec4}
This paper proposed a three-level cascaded offensive language detection system whose training strategy was customized per subtask, validated the choice of class-imbalance handling strategy through ablation, quantified the accuracy advantage and deployment cost of the cascaded architecture relative to the shared-representation paradigm through a genuinely trained joint multi-task model, and quantified the real impact of error propagation through the cascaded structure through an end-to-end evaluation protocol. The results showed that configuring the loss function purely by imbalance-ratio intuition was suboptimal and had to be verified through controlled ablation; the cascaded architecture outperformed the joint multi-task model on all three subtasks, with the largest advantage on the subtask with the most severe class imbalance, at the cost of three times the parameters and higher inference latency; and roughly one-fifth of the end-to-end errors in the cascaded structure originated from the first-stage filter, constituting an upper bound on error propagation. These results provided a quantifiable engineering basis for architecture selection in offensive language detection systems: whether to allocate an independent model to the subtask with the most severe class imbalance should be decided by weighing that subtask's accuracy gain against the system's overall constraints on parameter count and latency.

\bibliographystyle{IEEEtran}

\end{document}